\documentclass[letterpaper, 10pt, conference]{IEEEtrans}
\usepackage{graphicx}
\usepackage{hyperref}
\usepackage{amssymb}
\usepackage{pifont}
\usepackage{xcolor}
\usepackage{makecell}
\usepackage{cuted}
\usepackage{stfloats}
\usepackage{makecell}
\IEEEoverridecommandlockouts

\usepackage{graphics} % for pdf, bitmapped graphics files
\usepackage{epsfig} % for postscript graphics files
\usepackage{amsmath} % assumes amsmath package installed
\usepackage{amssymb}  % assumes amsmath package installed

\usepackage[utf8]{inputenc} % allow utf-8 input
\usepackage[T1]{fontenc}    % use 8-bit T1 fonts
\usepackage[table]{xcolor}
\usepackage[dvipsnames]{xcolor}
\usepackage{url}            % simple URL typesetting
\usepackage{amsfonts}       % blackboard math symbols
\usepackage{nicefrac}       % compact symbols for 1/2, etc.
\usepackage{microtype}      % microtypography
\usepackage{cleveref}
\usepackage{caption}
\usepackage{subcaption}

\usepackage[noadjust]{cite}
\usepackage[english]{babel}
\usepackage{hyperref}
\definecolor{darkblue}{RGB}{0,0,255}

\hypersetup{
    colorlinks=true,
    linkcolor=darkblue,
    filecolor=darkblue,
    urlcolor=darkblue,
    citecolor=darkblue
}

\usepackage[T1]{fontenc}            % For rendering/hyphenation of words w/ Latin characters
\usepackage{graphicx}               % figures & graphics
\usepackage{hyperref}               % link styles & signature
\usepackage{inconsolata}            % better texttt -- not sure if legal under IEEE?
\usepackage{nicefrac}               % compact symbols for 1/2, etc.

\usepackage[numbers,sort&compress]{natbib}        % better citations...

\newcommand{\cmark}{\textcolor{green}{\ding{51}}}
\newcommand{\xmark}{\textcolor{red}{\ding{55}}}
\newcommand{\name}{TeleDex}
\newcommand{\Name}{TeleDex}

\definecolor{red}{RGB}{150,31,1}
\definecolor{green}{RGB}{42,150,60}
\definecolor{blue}{RGB}{8,105,150}

\title{
\vspace{0.5cm}
\LARGE
\textbf{TeleDex: Accessible Dexterous Teleoperation}}

\author{
Omar Rayyan \hspace{1em} Maximilian Gilles$^{*}$ \hspace{1em} Yuchen Cui\\
University of California, Los Angeles\\
\thanks{$^{*}$Work done while visiting University of California, Los Angeles.}
}

\begin{document}
\maketitle
x\vspace{-1.2em}
\refstepcounter{figure}
\begin{strip}
\vspace{-6em}
\begin{center}
\includegraphics[width=0.99\textwidth]{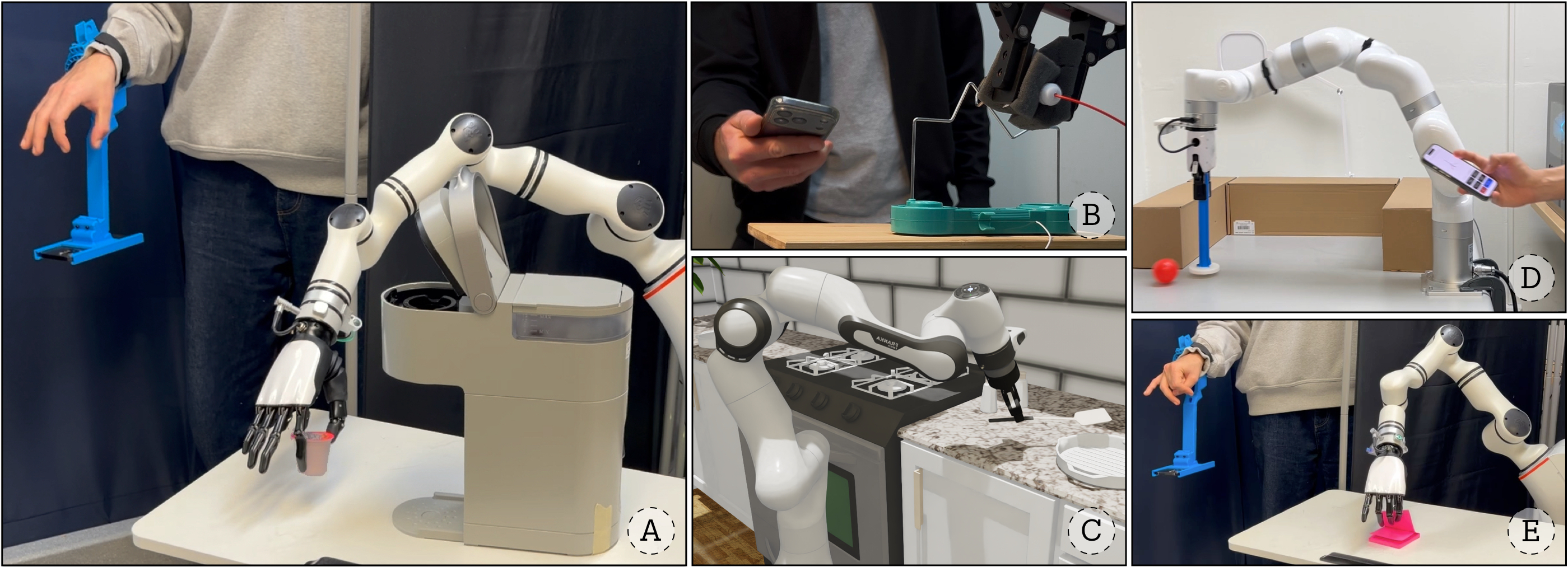}
\end{center}
\vspace{-0.6em}
{\small
\noindent\textbf{Fig. 1.} \textbf{TeleDex} enables accessible dexterous teleoperation using commodity smartphones.
The system streams 6-DoF wrist poses and 21-DoF articulated hand states to control robot arms and multi-fingered hands without external tracking or calibration.
The figure shows representative tasks performed using TeleDex: (A) espresso pod insertion into a coffee machine,
(B) precise manipulation through a buzz-wire game,
(C) DROID teleoperation in a MolmoSpaces kitchen within IsaacSim,
(D) a reactive goalkeeping task,
(E) a sticky-note peeling task.
}
\vspace{-0.8em}
\end{strip}

% \begin{abstract}
% \name\ is an open-source system that enables the teleoperation of dexterous hands and robot end-effectors in simulation and the real world using commodity smartphones. The system allows the streaming of 6-DoF device poses as well hand state estimates from iOS devices and maps them to appropriate robot control frames. \name\ supports both a hand-held phone-only mode and an optional 3D-printable hardware interface that extends its usability to dexterous robotic hands. The system is lightweight, requires no external tracking or calibration, and can provide haptic feedback to the operator. We demonstrate its utility across manipulation tasks that require dexterous control as well as reactive, low-latency responses. The project page is available at \href{https://orayyan.com/teledex}{orayyan.com/teledex}.
% \end{abstract}

\begin{abstract}

Despite increasing dataset scale and model capacity, robot manipulation policies still struggle to generalize beyond their training distributions. As a result, deploying state-of-the-art policies in new environments, tasks, or robot embodiments often requires collecting additional demonstrations. Enabling this in real-world deployment settings requires tools that allow users to collect demonstrations quickly, affordably, and with minimal setup. 
We present \Name{}, an open-source system for intuitive teleoperation of dexterous hands and robotic manipulators using any readily available phone. The system streams low-latency 6-DoF wrist poses and articulated 21-DoF hand state estimates from the phone, which are retargeted to robot arms and multi-fingered hands without requiring external tracking infrastructure. \Name{} supports both a handheld phone-only mode and an optional 3D-printable hand-mounted interface for finger-level teleoperation.
By lowering the hardware and setup barriers to dexterous teleoperation, \Name{} enables users to quickly collect demonstrations during deployment to support policy fine-tuning. We evaluate the system across simulation and real-world manipulation tasks, demonstrating its effectiveness as a unified scalable interface for robot teleoperation. All software and hardware designs, along with demonstration videos, are open-source and available at \href{https://www.orayyan.com/teledex}{orayyan.com/teledex}.

\end{abstract}

% Comment Max: General thoughts: i would make the point stronger and more clear what you mean by dexterous hand manipulation to highlight the difference to all the other hand held devices for data collection.
% Comment Max: So far we have differentiated from expensive tele op hardware, now What about umi etc? Can we differentiate here too? Maybe because we can teleop a dexterous hand? I would try to make the contribution here stronger wrt to pre-existing work.

\section{Introduction}

As recent improvements in robot learning have increasingly been driven by scaling data \cite{hu2024data,zhou2025vision,o2024open}, the ability to efficiently and accessibly collect demonstrations has become more important than ever. This need is amplified by limited model generalization, which often necessitates task-, scene-, and especially embodiment-specific data in deployment settings. Despite state-of-the-art open-sourced policies such as $\pi$~\cite{black2024pi_0,intelligence2025pi_}, GR00T~\cite{bjorck2025gr00t} and DreamZero \cite{ye2026worldactionmodelszeroshot} being trained on large and diverse datasets, they still require embodiment-specific fine-tuning when transferred to new robot platforms. In practice, this shifts the burden of data collection to end users, who must gather new demonstrations as tasks, environments, and embodiments change. Yet most existing manipulation data-collection interfaces rely on expensive, non-portable hardware \cite{yang2024ace, qin2023anyteleop, fu2024mobile}, limiting scalable deployment outside controlled laboratory environments. This is even more limiting for dexterous manipulation, where data collection is fundamentally harder and relies on less accessible hardware that add a lot of constraints. 

We believe modern phones represent an overlooked opportunity for robot manipulation data collection. They enable accurate, low-latency 6-DoF pose streaming and can act as  a general-purpose interface for robot manipulation data collection, particularly for tasks requiring fine, dexterous control without any external tracking infrastructure.

In this paper, we present \Name{}, an open-source system that uses phones as an accessible and intuitive interface for robot teleoperation and data collection. The system streams device pose and human hand state estimates and retargets them to robot end-effectors and dexterous grippers in both simulation and real-world settings. It is lightweight, requires no external tracking, and can be deployed with minimal setup across different arms and hands.

% \begin{figure*}[t]
% \centering
% \vspace{-0.8em}
% \includegraphics[width=0.98\linewidth]{figures/tea.jpg}
% \vspace{0.4em}
% \caption{\textbf{TeleDex enables accessible dexterous teleoperation using commodity phones.} The system streams 6-DoF wrist pose and 21-DoF hand states to control robot arms and multi-fingered hands without external tracking or calibration.}
% \vspace{-0.2em}
% \end{figure*}

Our contributions are as following:
\begin{itemize}
    \item An open-source lightweight teleoperation system consisting of a Python library and a publicly available iOS application for quick robot teleoperation and data collection.
    \item Support for dexterous hand teleoperation via a 3D-printable, wrist-mounted interface, integrated with Dex-Retarget~\cite{qin2023anyteleop} to support diverse robotic hand kinematics specified by URDFs.
    \item Native integration with MuJoCo~\cite{todorov2012mujoco}, enabling direct frame mapping to arbitrary MJCF models.
\end{itemize}

% 1. describe the need/problem: we need good ways to scale up data collection for robot manipulation, existing teleoperation methods often require special devices or not accurate  \\
% 2. motivation for this work: smartphones nowadays have advanced pose tracking sensors for augmented reality applications \\
% 3. contributions:\\
% - the app - read pose from iOS AR kit + UI for teleop \& data collection\\
% - dexterous version: finger detection with mediapipe\\
% - validated on across multiple sim benchmarks and real robots\\
% - validated policies collected from teleop

\section{Related Works}

\subsection{Scaling Robotics Data}

Recent work shows that imitation learning performance in robotics improves predictably as the amount of demonstration data increases \cite{hu2024data}, mirroring scaling trends observed in vision and language model. However, unlike web-scale domains, robotics lacks a large corpus of interaction data. 

To gather such data, a range of teleoperation systems are used like VR controllers and 3D-spacemouse devices (see Table~\ref{tab:device_comparison}). Other systems, such as GELLO~\cite{wu2023gello} and ALOHA~\cite{fu2024mobile}, use leader–follower setups with kinematically aligned hardware controllers to improve intuitiveness. While effective, these approaches require additional robot-specific hardware, increasing cost and limiting applicability at end user deployment setting. Moreover, these devices are limited to non multi-finger dexterous teleoperation.

On the other hand, handheld gripper systems such as UMI~\cite{chi2024universal} and CAP~\cite{cui2026contact} have recently emerged as robot-free data collection devices that abstract embodiment through an observation space aligned with deployment. However, demonstrations collected without an actual robot do not fully capture real robotic constraints. Additionally, the observation space of such devices is typically limited to wrist-mounted cameras, restricting their ability to capture the broader scene context typically available during teleoperation. %Besides, they lack multi-fingered hands.
Besides, adapting these interfaces to multi-fingered hands is non-trivial, requiring complex hardware modifications and additional post-processing~\cite{dexumi}.

\subsection{Phone-based Teleoperation Systems}

Commodity devices can serve as effective teleoperation devices as well (see Table~\ref{tab:device_comparison}).
Roboturk~\cite{mandlekar2018roboturk} was the first to demonstrate that phones can be used as low-cost teleoperation devices for large-scale robot data collection by streaming their 6-DoF end-effector commands in simulation. Teleoperation stacks, such as TeleMOMA~\cite{dass2024telemoma}, have since adopted phone-based interfaces. However, these approaches are limited to 6-DoF end-effector pose control and do not allow the transmission of articulated finger-level commands, making them unsuitable for dexterous manipulation tasks that require fine-grained control, despite their portability and ease of deployment. In addition, neither of these systems are distributed as publicly, limiting their adoption.

Recent work further validates the suitability of phones as teleoperation interfaces. User studies in COBALT~\cite{agarwal2025cobalt} show that phone-based control can outperform traditional teleoperation hardware in efficiency and usability, reinforcing the promise of commodity devices for scalable data collection. These findings motivate extending phone-based teleoperation beyond end-effector control to support high-DoF dexterous manipulation.

% \begin{table}[t]
% \centering
% \scriptsize
% \renewcommand{\arraystretch}{1.3}
% \caption{Comparison of Teleoperation Interfaces for Demonstration Collection. \omar{can someone verify the bins}}
% \label{tab:device_comparison}
% \begin{tabular}{lcccc}
% \hline
% \textbf{System} 
% & \makecell{\textbf{Commodity} \\ \textbf{Device}}
% & \makecell{\textbf{Fine} \\ \textbf{Control}}
% & \makecell{\textbf{No} \\ \textbf{Calibration}}
% & \makecell{\textbf{Multi-Finger} \\ \textbf{Control}} \\
% \hline
% Keyboard         & \cmark & \xmark & \cmark & \xmark \\
% VR               & \xmark & \cmark & \cmark & \xmark \\
% 3D Mouse         & \xmark & \xmark & \cmark & \xmark \\
% GELLO            & \xmark & \cmark & \cmark & \xmark \\
% UMI              & \xmark & \cmark & \xmark & \xmark \\
% ACE-Teleop       & \xmark & \cmark & \cmark & \cmark \\
% AnyTeleop        & \cmark & \xmark & \cmark & \cmark \\
% COBALT        & \cmark & \cmark & \cmark & \xmark \\
% \textbf{TeleDex} & \cmark & \cmark & \cmark & \cmark \\
% \hline
% \end{tabular}
% \renewcommand{\arraystretch}{1.0}
% \end{table}
% \normalsize

\begin{table}[t]
\centering
\scriptsize
\renewcommand{\arraystretch}{1.3}
\caption{Comparison of Common Teleoperation Interfaces for Collecting Human Demonstration Data.}
\label{tab:device_comparison}
\begin{tabular}{lcccc}
\hline
\textbf{System} 
& \makecell{\textbf{Off-the-Shelf} \\ \textbf{Device}}
& \makecell{\textbf{Fine} \\ \textbf{Control}}
& \makecell{\textbf{No} \\ \textbf{Calibration}}
& \makecell{\textbf{Multi-Finger} \\ \textbf{Control}} \\
\hline
Keyboard         & \cmark & \xmark & \cmark & \xmark \\
VR               & \cmark & \cmark & \cmark & \xmark \\
SpaceMouse         & \cmark & \cmark & \cmark & \xmark \\
GELLO            & \xmark & \cmark & \cmark & \xmark \\
UMI              & \xmark & \cmark & \xmark & \xmark \\
ACE-Teleop       & \xmark & \cmark & \cmark & \cmark \\
AnyTeleop        & \cmark & \xmark & \cmark & \cmark \\
COBALT           & \cmark & \cmark & \cmark & \xmark \\
\rowcolor[gray]{.9} \textbf{TeleDex} & \cmark & \cmark & \cmark & \cmark \\
\hline
\end{tabular}
\renewcommand{\arraystretch}{1.0}
\end{table}

\subsection{Dexterous Hand Teleoperation}

Dexterous hand teleoperation for multi-finger control has been explored through both vision-based and hardware-based approaches (see Table~\ref{tab:device_comparison}). AnyTeleop~\cite{qin2023anyteleop} enables dexterous manipulation using vision-based hand pose estimation combined with a retargeting pipeline. The system decouples arm and hand control by mapping human wrist pose to the robot arm and articulated finger motion to multi-fingered robotic hands via Dex-Retarget. While effective, such approaches rely on camera-based hand pose estimation from a fixed external viewpoint, making them sensitive to occlusions, viewpoint changes, and depth ambiguity as outlined in their limitations. Moreover, wrist pose is inferred from image-based depth cues rather than direct 6-DoF sensing, which can introduce noise during fast motions and contact-rich interactions.
Other systems like ACE-Teleop \cite{yang2024ace} use dedicated sensing hardware and external tracking systems to capture hand motion. While these setups can provide accurate articulated tracking, they require additional sensors, and non-portable hardware.

% \begin{figure*}[t!]
%   \centering
%     \includegraphics[width=\textwidth]{figures/server.pdf}

%   \caption{Overview of \name{}.}
%   \label{fig:system_overview}
% \end{figure*}

\begin{figure}[]
    \centering
    \includegraphics[width=0.98\columnwidth]{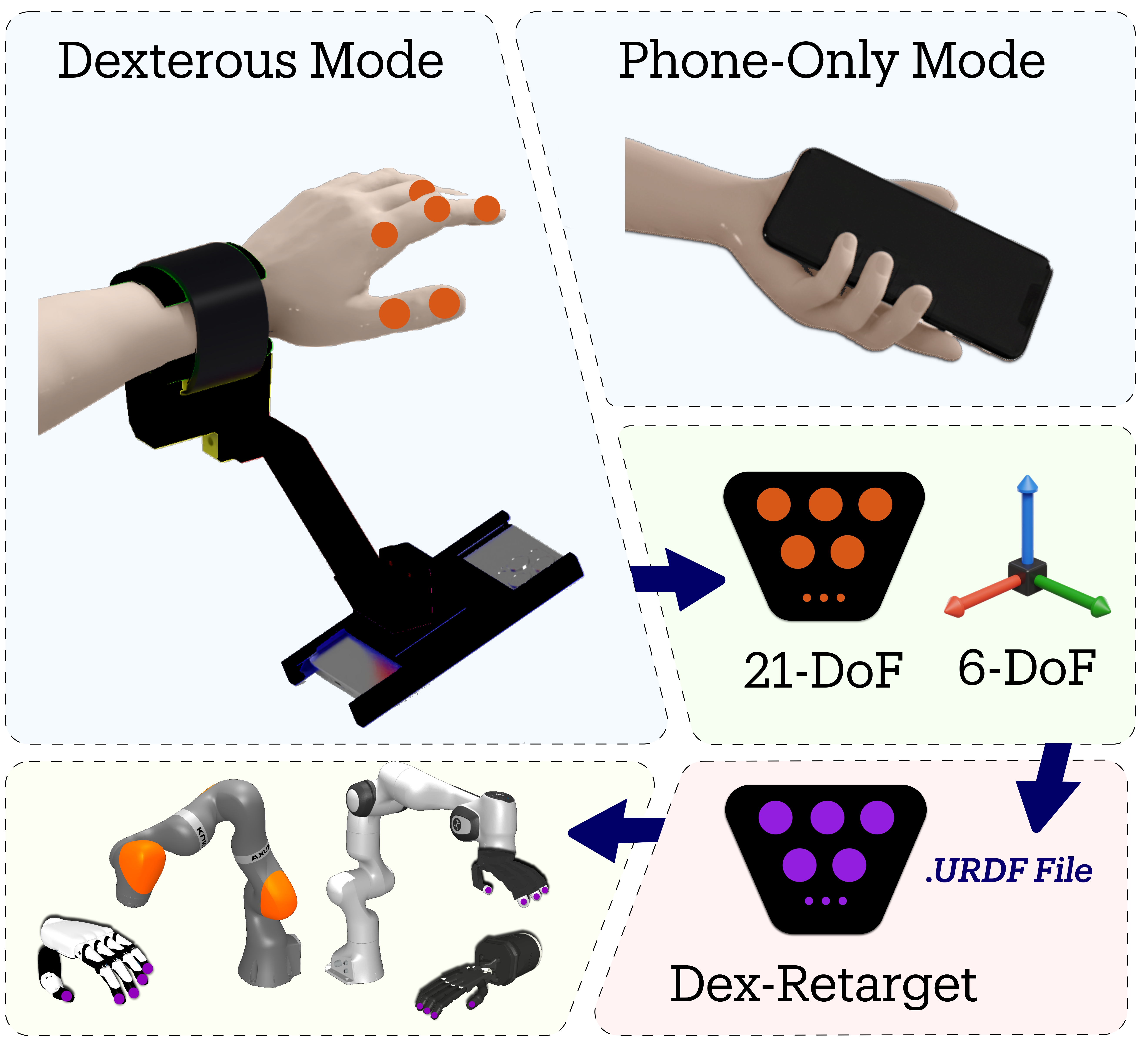}
    \caption{\textbf{TeleDex System Architecture.} The system supports two settings: (left) a \textbf{Dexterous Setting} using a 3D-printable wrist-mounted interface for 21-DoF hand and 6-DoF wrist tracking, and (right) a \textbf{Phone-Only Setting} for 6-DoF end-effector control. Captured states are processed via Dex-Retarget to map inputs to arbitrary robotic URDF models.}
    \label{fig:hardware}
\end{figure}

\section{System Overview}

\Name{} supports two teleoperation settings (see \cref{fig:hardware}). 
(A) In the basic handheld setting, a single smartphone acts as a 6-DoF input device for Cartesian end-effector control. 
(B) For tasks requiring finer manipulation, we provide an optional 3D-printable wrist-mounted interface inspired by Ace-Teleop~\cite{yang2024ace} that enables articulated multi-finger tracking and retargeting to multi-DoF robotic hands.

\subsection{Basic Hand-held Setting}
In the handheld configuration, the phone acts as a self-contained 6-DoF tracking device. ARKit\footnote{https://developer.apple.com/augmented-reality/arkit} estimates the device pose without requiring additional hardware or calibration. The user simply holds the phone and uses its motion as a 6-DoF input for controlling the robot end-effector.

On the robot side, at the start of teleoperation the system is reset by recording the robot's initial end-effector pose and the current phone pose as a reference. Let ${}^{A}T_{P}^{0}$ denote the phone pose at reset and ${}^{A}T_{P}$ the current phone pose, both expressed in the reference frame $\{A\}$. The relative phone motion is $$\Delta {}^{A}T_{P} = \left({}^{A}T_{P}^{0}\right)^{-1} {}^{A}T_{P}$$
A fixed alignment transform ${}^{P}T_{R}$ is used to account for the coordinate offset between the phone frame and the robot control frame. The commanded robot pose is then computed as 
$${}^{R}T_{EE} = {}^{R}T_{EE}^{0} \, {}^{P}T_{R} \, \Delta {}^{A}T_{P}$$ 
We provide the option to scale the position changes to vary the sensitivity based on the need of the user.

\subsection{Wrist-Mounted Setting}

To support finger-level manipulation, we open-source 3D-printable arm-mounted interface. We provide multiple sizes of adapters to accommodate varying wrist circumferences and phone dimensions. The components are assembled using screws. A double-sided hook-and-loop strap is used to attach it to the wrist.

In this configuration, the phone is rigidly attached to the operator’s hand with a fixed offset, allowing the system to jointly capture articulated hand landmark estimates using the front camera while using ARKit to track the phone's pose. Unlike external camera-based approaches, where the user must ensure their hand remains within the field of view of a fixed camera since the pose is inferred from image-based depth cues, the arm-mounted configuration allows the operator to move freely because the camera moves together with the hand.

During teleoperation, as in the handheld setting, the device streams the phone pose ${}^{A}T_{P}$ to the control server. In addition, it also transmits the estimated 21 hand landmarks. Since the phone is rigidly attached to the operator’s arm, a fixed transform ${}^{P}T_{H}$ encodes the offset between the phone frame and the hand frame. The resulting hand pose is computed as ${}^{A}T_{H} = {}^{A}T_{P} {}^{P}T_{H}$ and is used to control the robot arm.

In parallel, the hand landmarks are retargeted to a multi-DoF robotic hand using Dex-Retarget~\cite{qin2023anyteleop} given a URDF model of the target hand. This decoupled design enables the same interface to generalize across robotic hands with different kinematic structures.

\subsection{TeleDex App and Python API}

\begin{figure}[h]
\centering
\includegraphics[width=0.9\columnwidth]{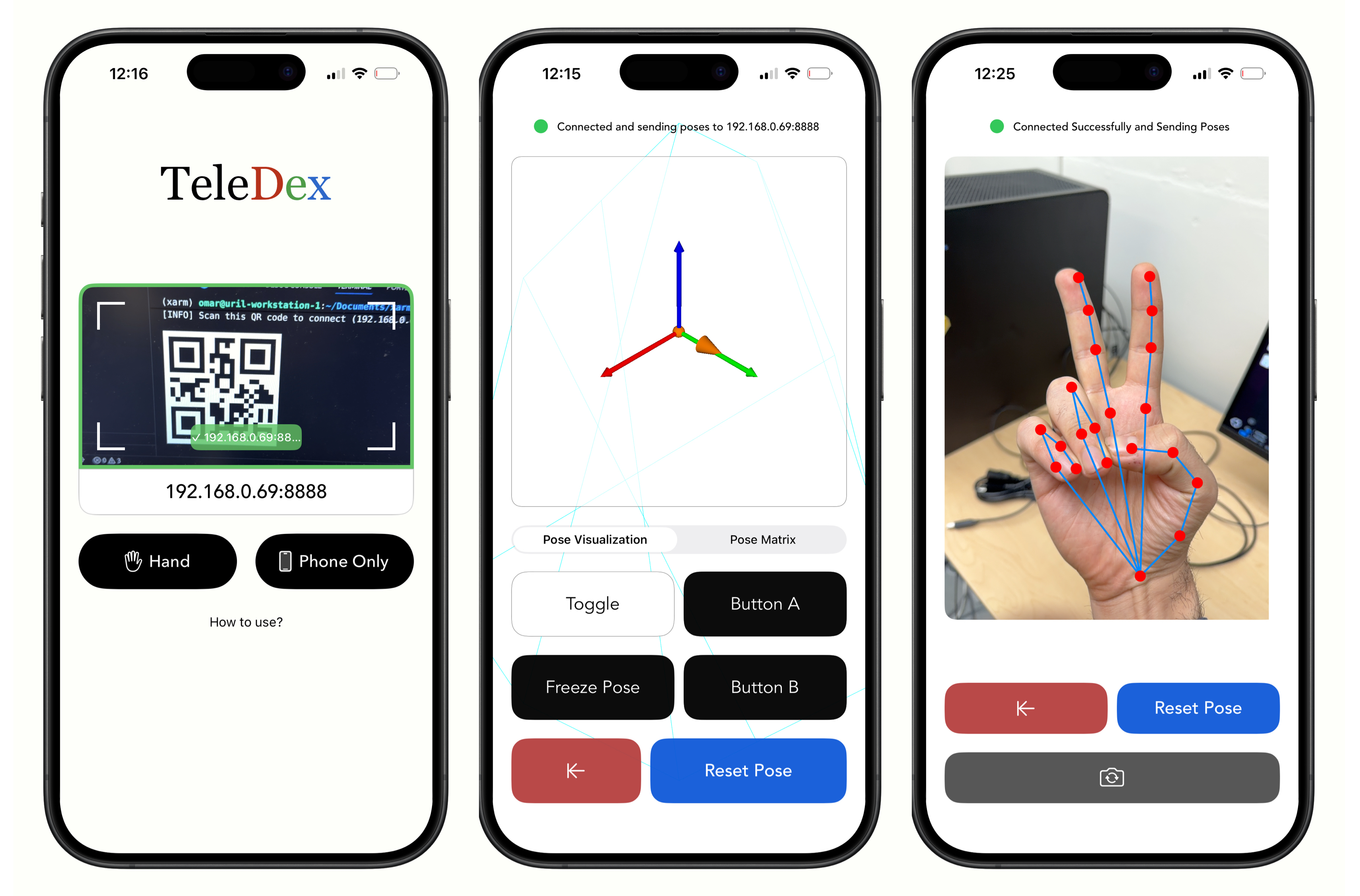}
\caption{Screenshots of TeleDex phone application.} % used during teleoperation.}
\label{fig:teledex_app}
\end{figure}

\Name{} is publicly distributed on the App Store\footnote{\url{https://apps.apple.com/us/app/teledex/id6612039501}} together with a companion Python package for integration with robotic systems. Screenshots of the app interface are shown in \cref{fig:teledex_app}. The \Name{} mobile application implements the teleoperation interface for both supported control settings. The robot-side server is first launched using the provided Python API. Once the robot and smartphone are connected to the same network, the server generates a code to be scasnned through the mobile app in order establish a connection. After the connection is confirmed, teleoperation can be enabled immediately, allowing real-time control and data streaming.

The app includes configurable on-screen buttons with user-defined callback functions, enabling flexible adaptation to task-specific requirements. For example, users can dynamically lock selected control dimensions, reset reference frames, or adjust motion scaling during operation. This design allows practitioners to tailor the interface to diverse manipulation scenarios without modifying the core application.

On the robot side, the \Name{} Python API provides a lightweight interface for receiving streamed pose and hand-tracking data, performing necessary coordinate frame transformations, and exposing the processed states to downstream control pipelines. The package is framework-agnostic and can be integrated with common robotics stacks. It is available via \texttt{pip install teledex}, facilitating straightforward deployment and reproducibility.

% \begin{figure}[h]
%     \centering
%     \includegraphics[width=0.7\columnwidth]{figures/cad_rendering.png}
%     \caption{
% The hand-mounted interface comprises four 3D-printable components: a wrist adapter, wrist mount, linking element, and phone mount. The wrist adapter and phone mount come in multiple sizes to accommodate varying wrist circumferences and phone dimensions. The components are assembled using screws. A double-sided hook-and-loop strap is used to attach it to the wrist.
%     }
%     \label{fig:hardware}
% \end{figure}

% \begin{figure}[t!]
%   \centering
%   \makebox[\linewidth][c]{%
%     \includegraphics[width=1.0\linewidth]{figures/examples.pdf}
% .  }
% \caption{Overview of \name{}.}
%   \label{fig:system_overview}
% \end{figure}

\section{Experimental Evaluation}
\label{sec:result}

\begin{figure}[b]%[h!]
\centering
\includegraphics[width=\linewidth]{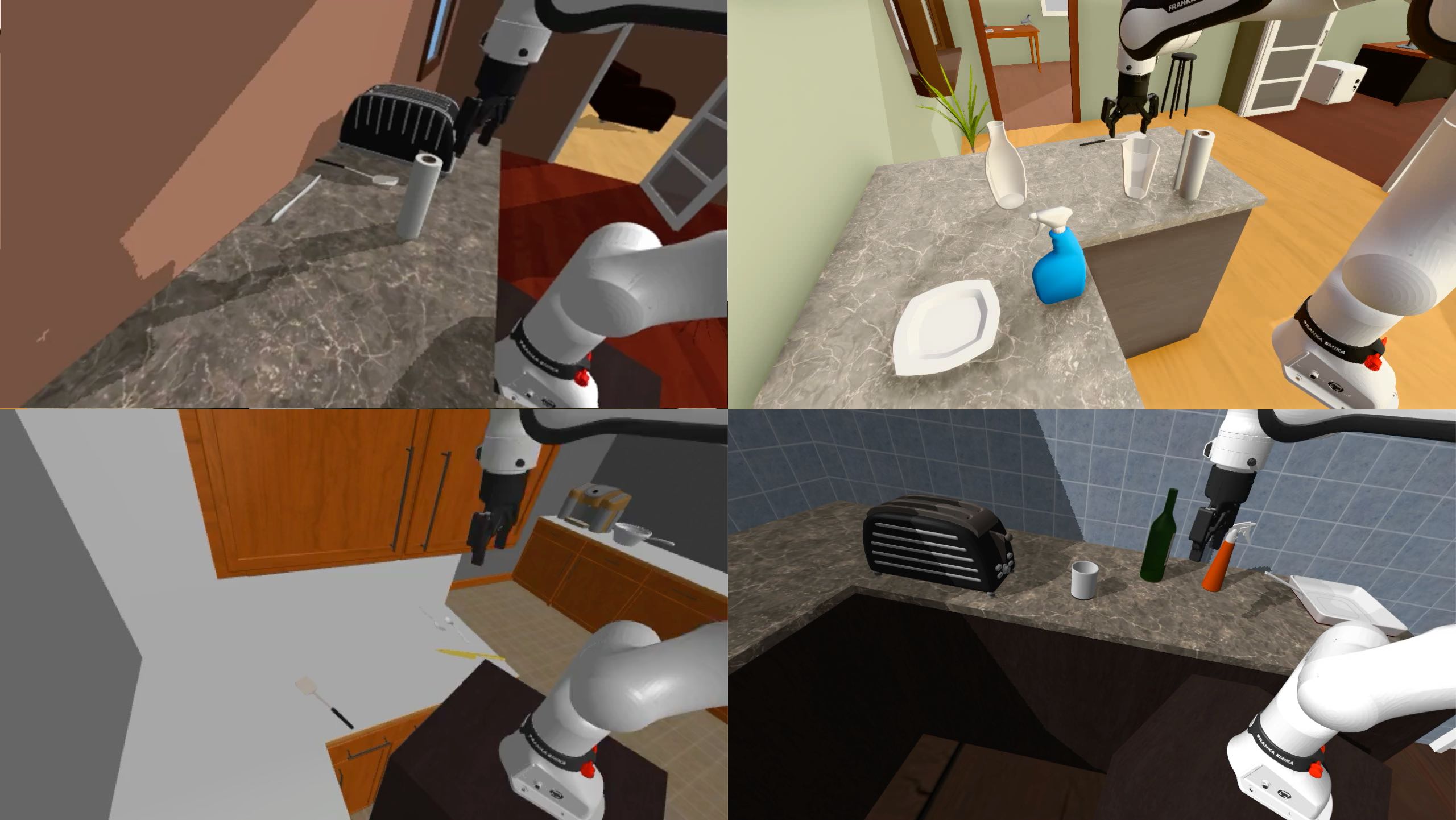}
\caption{
Example tasks from MolmoSpaces-Bench.
}
\label{fig:molmo_tasks}
\end{figure}

To systematically assess the practicality and effectiveness of \Name{}, we evaluate it with two primary research questions:

\begin{itemize}
    \item \textbf{RQ1:} How efficiently can \Name{} be used to collect demonstrations compared to commonly used commodity control interfaces?
    \item \textbf{RQ2:} Can \Name{} enable \textit{reliable} and \textit{intuitive} dexterous teleoperation across representative manipulation tasks? %, both qualitatively and quantitatively?
\end{itemize}

To address RQ1, we conduct a user study in simulation featuring a suite of teleoperation tasks and measure the amount of successful demonstrations that can be collected within a fixed time budget. To address RQ2, we perform a user study on a real-world dexterous robotic platform across representative manipulation tasks and evaluate user performance. In addition, we showcase a range of dynamic tasks enabled by \Name{} and demonstrate downstream policy learning in a simulated environment.

\subsection{Experimental Setup}

We evaluate \Name{} across three platforms:

\begin{itemize}
    \item \textbf{xArm7~\cite{ufactory_xarm7} (real-world):} A 7-DoF arm equipped with a parallel-jaw gripper for table-top teleoperation tasks.
    \item \textbf{RealMan~\cite{realman_robotics} (real-world):} A mobile manipulator equipped with a multi-finger dexterous hand~\cite{oymotion_hand} for articulated manipulation tasks.
    \item \textbf{MolmoSpaces-Bench (simulation):} A large-scale simulation benchmark \cite{kim2026molmospaces} where \Name{} is deployed as a teleoperation interface for collecting demonstrations.
\end{itemize}

All experiments uses an iPhone running the publicly available version of the \Name{} application.

\subsection{RQ1: Efficiency of Collecting Demonstrations}

We start by evaluating how efficiently \Name{} enables demonstration collection by comparing it to commonly used commodity control interfaces with a user study.

\subsubsection{Task Setup}
We consider two tasks: \textit{MS-Pick} and \textit{MS-Open}. For \textit{MS-Pick}, the user is instructed to pick a specified object in the scene and lift it up. Objects are placed on flat surfaces and surrounded by other items. For the \textit{MS-Open} task, the user is instructed to open a drawer (prismatic joint) or cabinet door (revolute joint) in the scene. %The drawer can be opened either through linear end-effector motions or through circular motions. 
% In total, we constructed 15 different experimental runs for each task.

\begin{figure}%[h!]
\centering
\includegraphics[width=0.48\columnwidth]{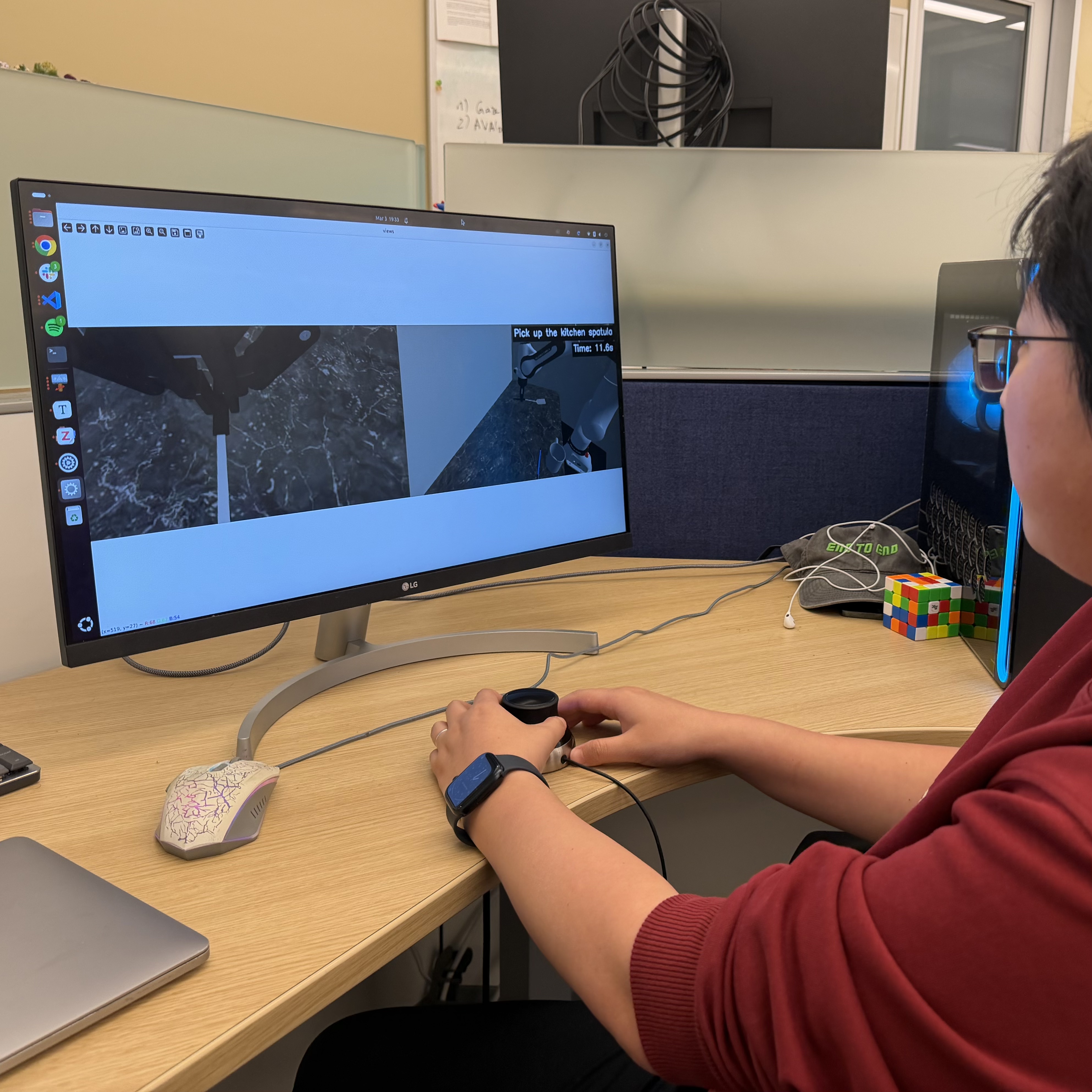}
\hfill
\includegraphics[width=0.48\columnwidth]{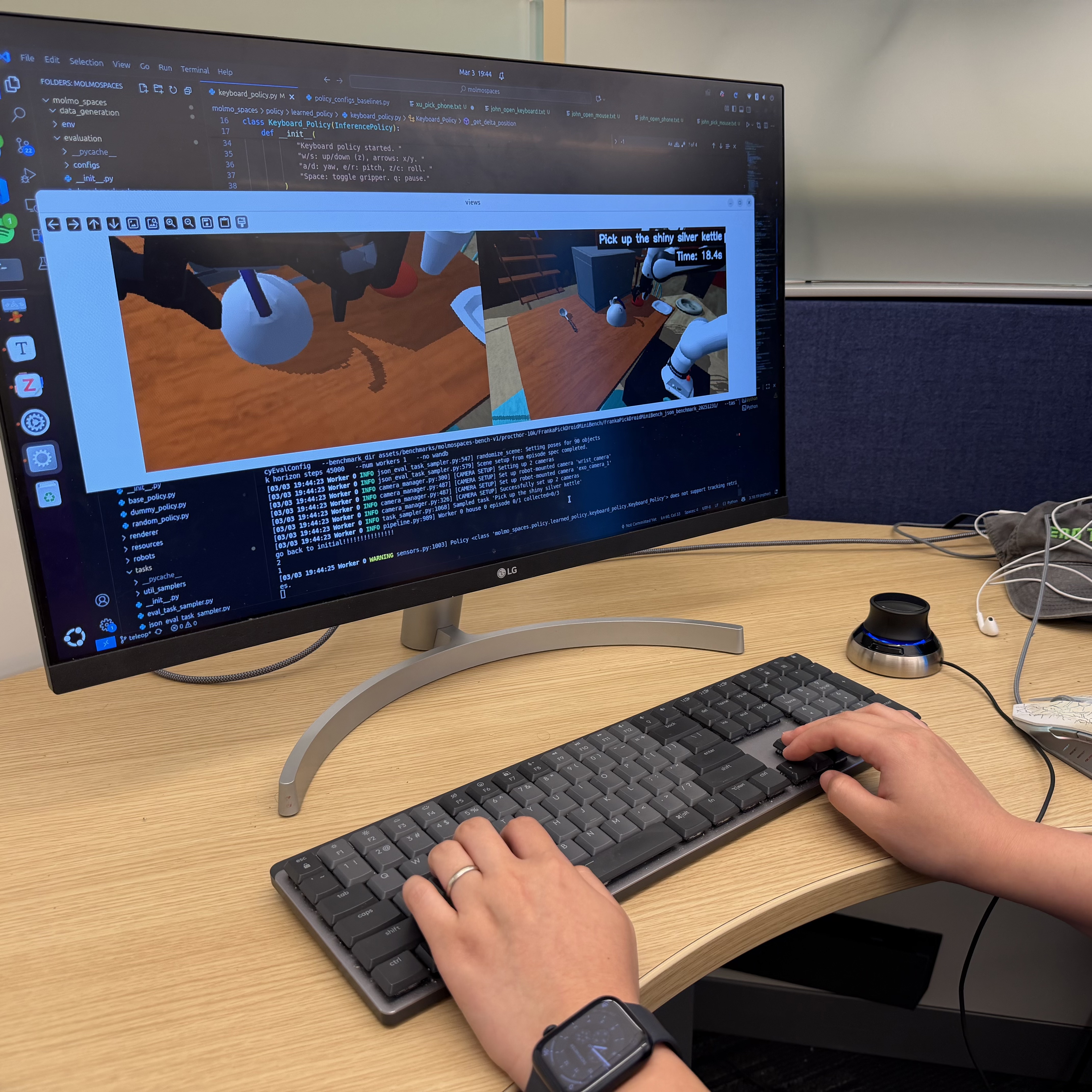}
\caption{
MolmoSpaces-Bench user study setup: left, using SpaceMouse; right, using a keyboard.
}
\label{fig:user_study}
\end{figure}
\subsubsection{Metrics}
For each task, we measure the average time taken per success. We set a time limit at 30\,s and consider a trial unsuccessful if it exceeds this time. A grasp is considered successful if the object is lifted more than 5\,cm, or if the drawer or cabinet is opened more than 15\% of its joint range.

\subsubsection{Participants and Procedures}
We recruited five participants. Each participant used all three devices and completed fifteen episodes per device, resulting in a total of 225 episodes. It is worth noting that all users had prior with the non-phone control interfaces; however, non had a prior experience with systems similar to \Name{}. 
Before the experiment, each participant was given a two-minute familiarization period for each interface. After this warm-up phase, participants performed 10 trials per task with each of the three control interfaces under evaluation: keyboard control, a 3Dconnexion SpaceMouse~\cite{3dconnexion_spacemouse}, and the proposed \Name{} interface. Each trial had a time limit of 30 seconds. The order of the teleoperation interfaces was randomized across participants to mitigate ordering effects.

% \begin{itemize}
% \item Keyboard control
% \item 3D SpaceMouse
% \item \Name{} phone interface
% \end{itemize}

\begin{figure}[t]
    \centering
    \includegraphics[width=0.89\columnwidth]{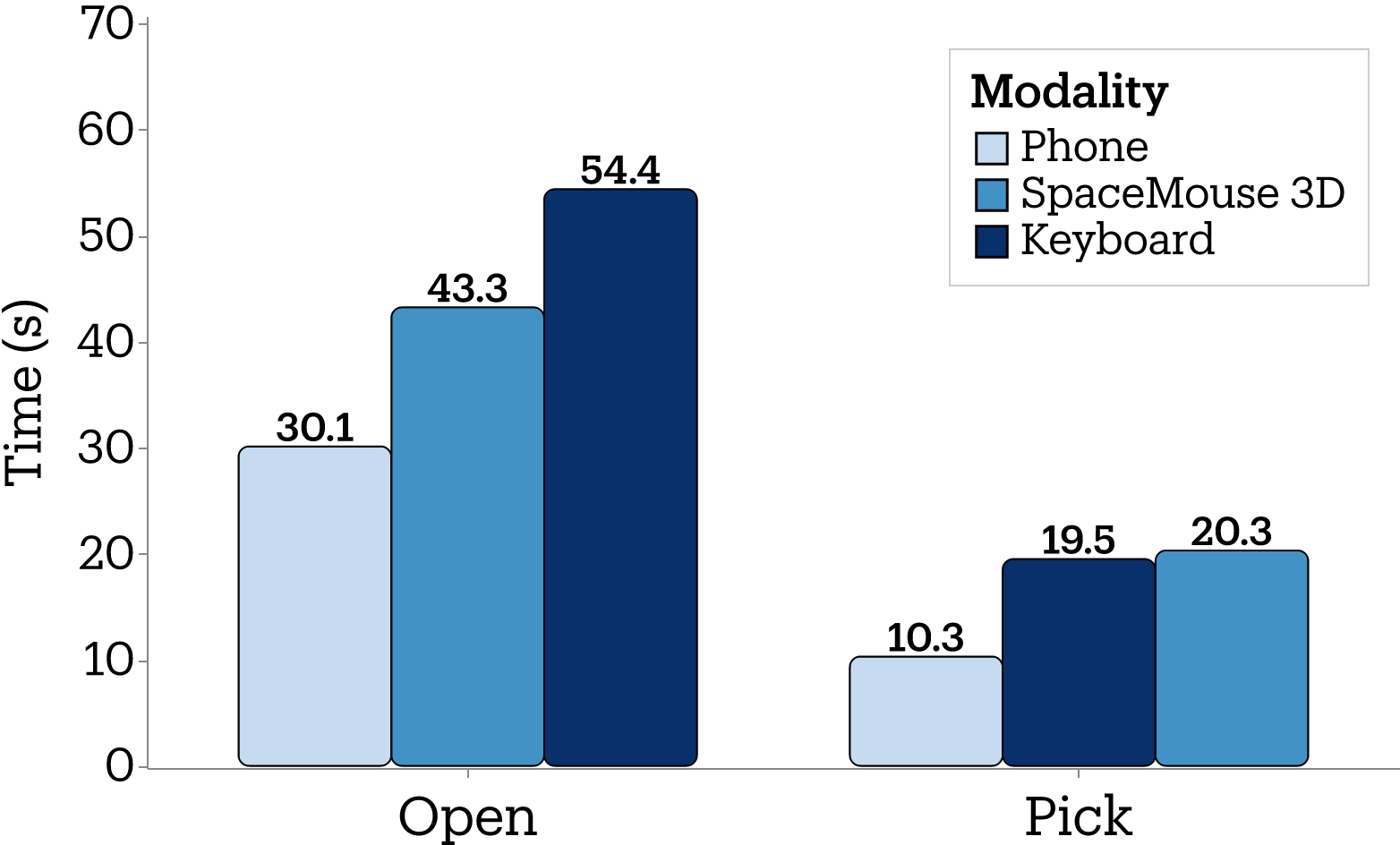}
    \caption{Average time-to-success for the \textit{MS-Open} and \textit{MS-Pick} tasks, across users in MolmoSpaces-Bench.}
    \label{fig:time_until_success}
\end{figure}

\subsubsection{Results}
Figure~\ref{fig:time_until_success} shows the average time required to obtain a successful demonstration for the \textit{MS-Open} and \textit{MS-Pick} tasks. Across both tasks, \Name{} enables faster and more efficient demonstration collection compared to the 3D SpaceMouse and keyboard baselines. While keyboard-based demonstrations are easy to instruct, users struggle in controlling different rotations. While SpaceMouse offers more intuitive rotation control, it requires time for users to adapt to it and remains difficult to control as quickly as the phone.

\subsection{RQ2: Intuitive Dexterous Teleoperation}

To evaluate the dexterous teleoperation capabilities of \Name{}, we conduct a user study where participants perform dexterous manipulation tasks using the RealMan~\cite{realman_robotics} platform. %Participants with no prior experience using our system are asked to performs all tasks after a 2 minutes familiarization period with the interface. 

\begin{figure*}[b]
\centering
\includegraphics[width=0.98\linewidth]{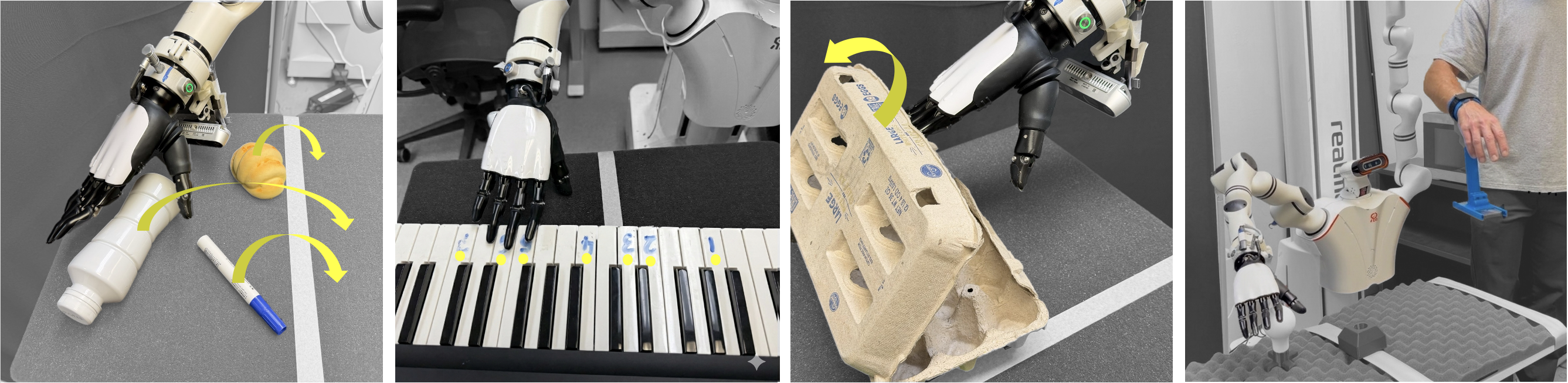}
\caption{
Dexterous tasks used for user study of RQ2: %quantitative and qualitative evaluation.
\textbf{(a)} pick-and-place with objects of varying geometries.
\textbf{(b)} targeted piano key pressing.
\textbf{(c)} egg container opening and object extraction.
\textbf{(d)} participant performing tasks during the user study.
}
\label{fig:dex_tasks}
\end{figure*}

\subsubsection{Task Description}

We evaluate a set of representative dexterous manipulation tasks.

For quantitative evaluation we consider three tasks as shown in Fig~\ref{fig:dex_tasks}. All these tasks are designed to capture precision grasping, finger articulation, and contact-sensitive manipulation. 

\begin{itemize}
\item \textbf{Pick-and-Place (Multi-Shape):} Grasping and placing three objects in two-ways (six total pick-places) required the ability to handle varying shapes and arbitrary orientations.
\item \textbf{Targeted Piano Key Pressing:} Pressing specified keys in a piano requires independent finger articulation and fine-grained hand control.
\item \textbf{Egg Container Opening and Picking:} Opening a hinged egg carton requires fine-grained control.
\end{itemize}

For the \textit{pick-and-place task}, three objects of varying size and shape (a bottle, a marker, and a soft baguette) were placed on a flat foam surface. The user is instructed to pick up each object, move it to the opposite side, and then return it to its original side.
For the \textit{targeted piano key pressing task}, the user is instructed to press marked keys in a specified sequence.
For the \textit{egg container opening task}, the goal is to open the container, grasp the object inside, and place it next to the container.

\subsubsection{Metrics}
For each task we report the mean completion time across successful trials.  We compare TeleDex teleoperation performance against an oracle(human) performance, where the same tasks are performed directly by a human at a speed representative of an intent to demonstrate and teach the task. This baseline approximates the speed at which demonstrations could realistically be produced for policy training.

\subsubsection{Participants and Procedures}
We recruited a total of 7 participants with no prior experience using our system. Participants were asked to perform the target tasks after a two-minute familiarization period.

\subsubsection{Results}

Table~\ref{tab:dex_time} reports the mean completion time across successful trials. All participants were able to complete all tasks using \Name{} after the short familiarization period. The interface enabled users to perform various grasping styles and coordinated finger motions required for precise manipulation tasks such as targeted key pressing and container opening.

\begin{table}[h]
\centering
\caption{Completion Time (s) on Successful Trials (mean$\pm$std)}
\label{tab:dex_time}
\begin{tabular}{lccc}
\hline
Method & \makecell{6 Pick \\ and Place} & \makecell{Piano \\ Keys Press} & \makecell{Eggs \\ Container} \\
\hline
\makecell[l]{Human \\ (policy-aware speed)} & 23 & 21 & 7 \\
TeleDex & $74.0 \pm 39.0$ & $75.8 \pm 19.7$ & $18.7 \pm 6.6$ \\
\hline
\end{tabular}
\end{table}
%TBD

% \subsection{Additional Dynamic Experiments}
% We additionally evaluate TeleDex in dynamic manipulation scenarios \textit{catching}, \textit{goalkeeping}\textit{buzz-wire loop}, as shown in Fig~\ref{fig:reactive} requiring quick and precise spatial control. These demonstrations illustrate the responsiveness of the interface during fast interactions and contact-sensitive manipulation. Operators were able to complete the tasks and complete movements adjustments without observable tracking drift.
% We refer to our additional videos on our project page to get a feeling how reactive (\textit{object catching} and \textit{ goalkeeping task}) and fine-grained control (\textit{buzz-wire loop} \Name{} allows.
\subsection{Complex Manipulation Experiments}
\begin{figure}[t]
\centering
\includegraphics[width=\columnwidth]{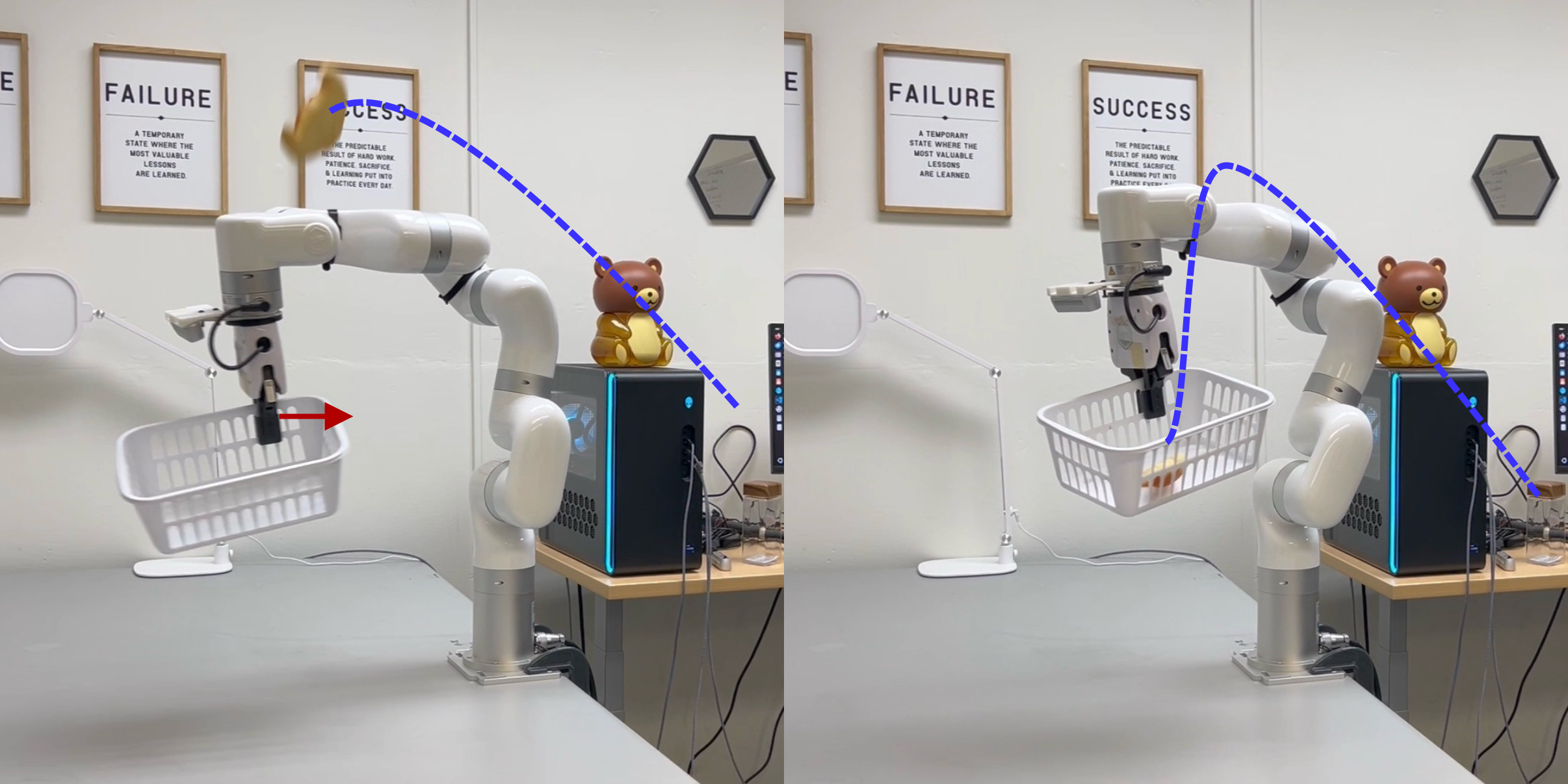}

\vspace{0.3em}

\includegraphics[width=0.48\columnwidth]{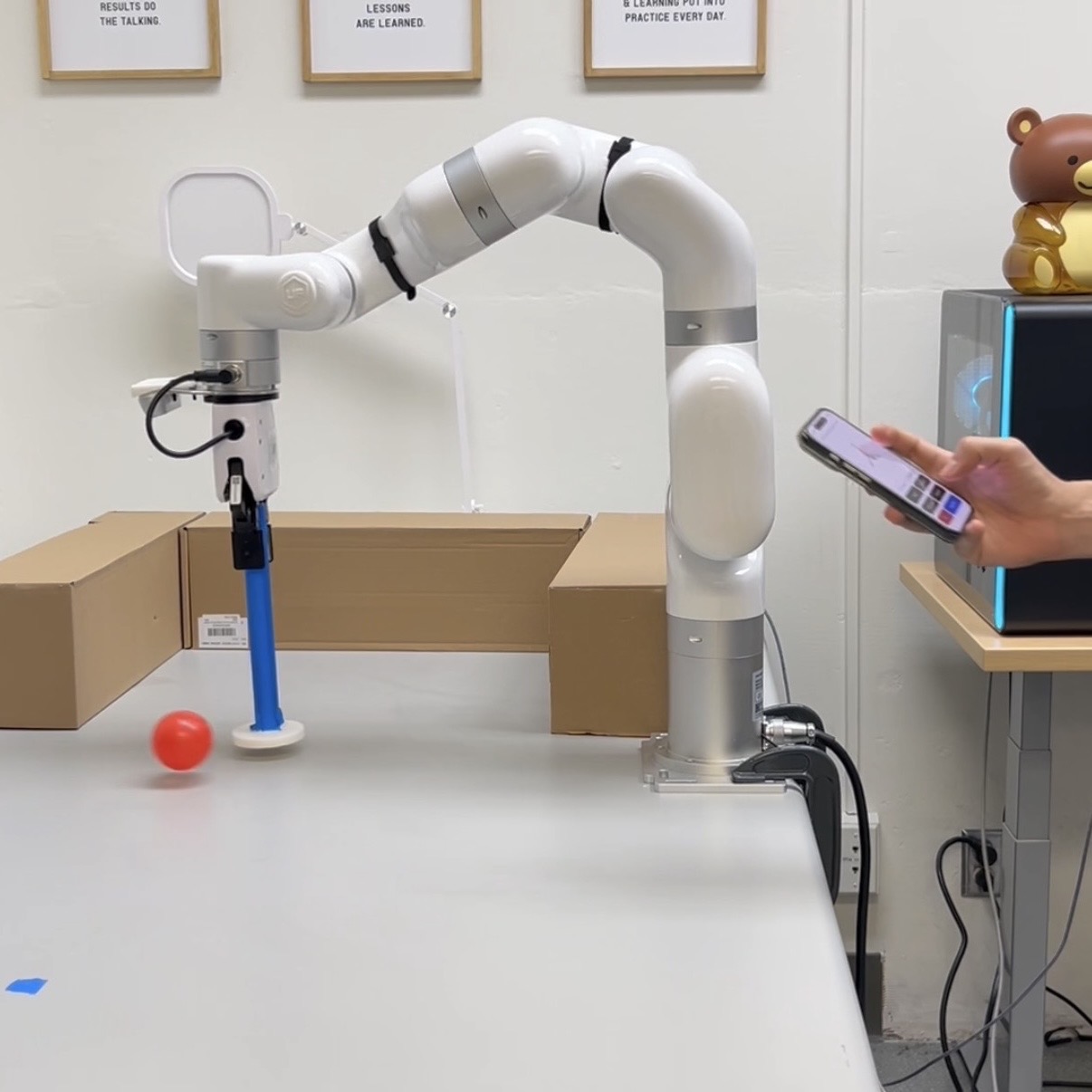}
\hfill
\includegraphics[width=0.48\columnwidth]{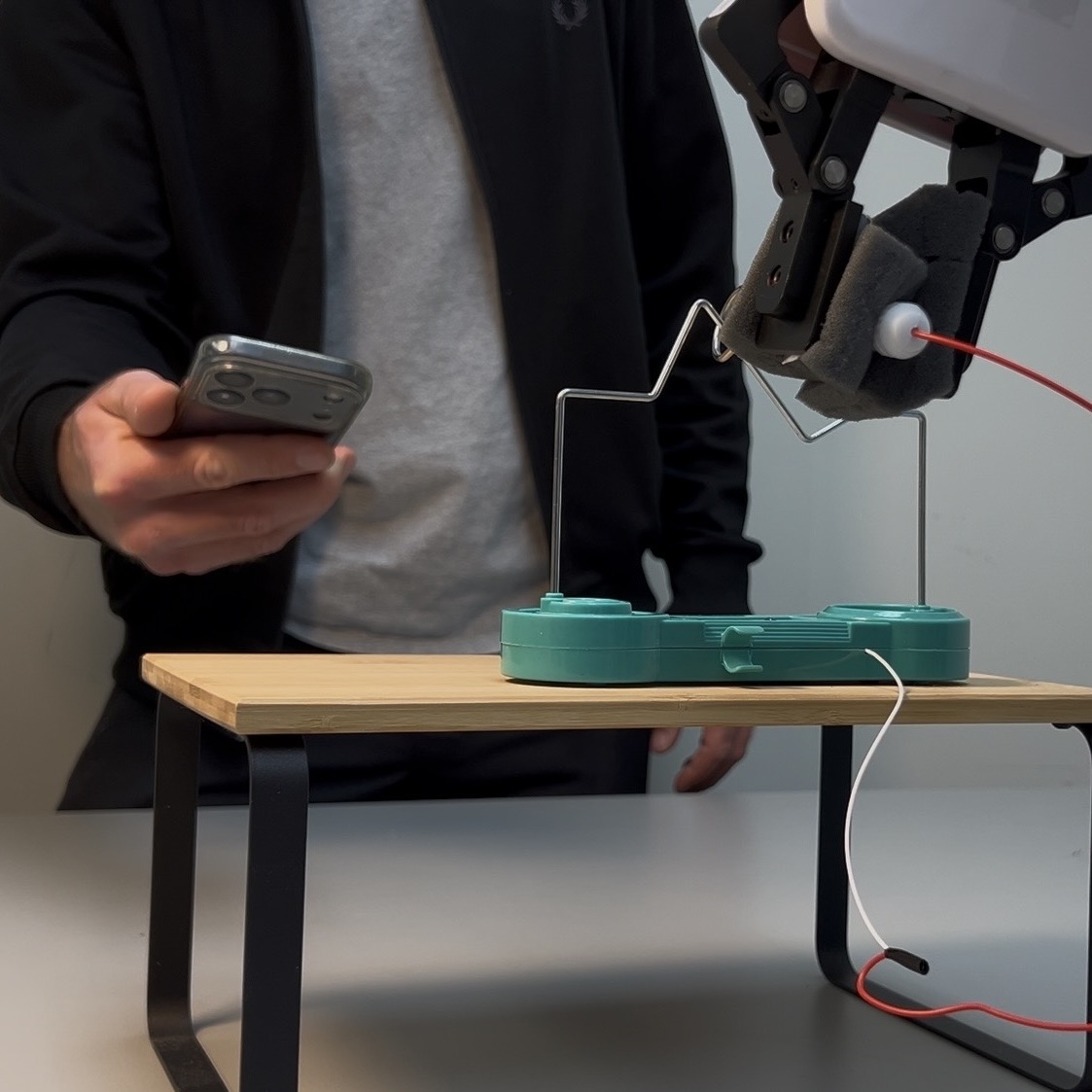}

\caption{
Demonstration of dynamic, reactive and precise manipulation examples with XArm7 and \Name:
(a) dynamic object catching,
(b) reactive goalkeeping,
(c) precise buzz-wire loop tracing.
}
\label{fig:reactive}
\end{figure}

We further evaluate \Name{} in a set of challenging manipulation scenarios, including dynamic \textit{object catching}, reactive \textit{goalkeeping}, and precise \textit{buzz-wire loop} tracing, as illustrated in Fig.~\ref{fig:reactive}. These tasks demand rapid response times, accurate spatial coordination, and stable contact-aware control, thereby stress-testing both the responsiveness and precision of the teleoperation interface.

Across all scenarios, operators were able to successfully complete the tasks via performing fine-grained corrective motions in real time, without observable tracking drift or loss of control. In particular, the dynamic tasks (\textit{catching} and \textit{goalkeeping}) highlight the low-latency and reactive characteristics of the system, whereas the \textit{buzz-wire loop} task emphasizes its capability for steady, high-precision manipulation under tight spatial constraints. For these tasks, the provided Python API of \Name{} and the customizable buttons in the app support locking control dimensions and dynamically scaling motion sensitivity, enabling task-specific adaptation of the control behavior.

Additional demonstration videos are provided on the project webpage, showcasing the reactivity of \Name{} in dynamic interactions as well as its fine-grained control performance in contact-sensitive manipulation tasks.

\subsection{Policy Training}

\begin{figure}[h]
\centering
\includegraphics[width=0.95\columnwidth]{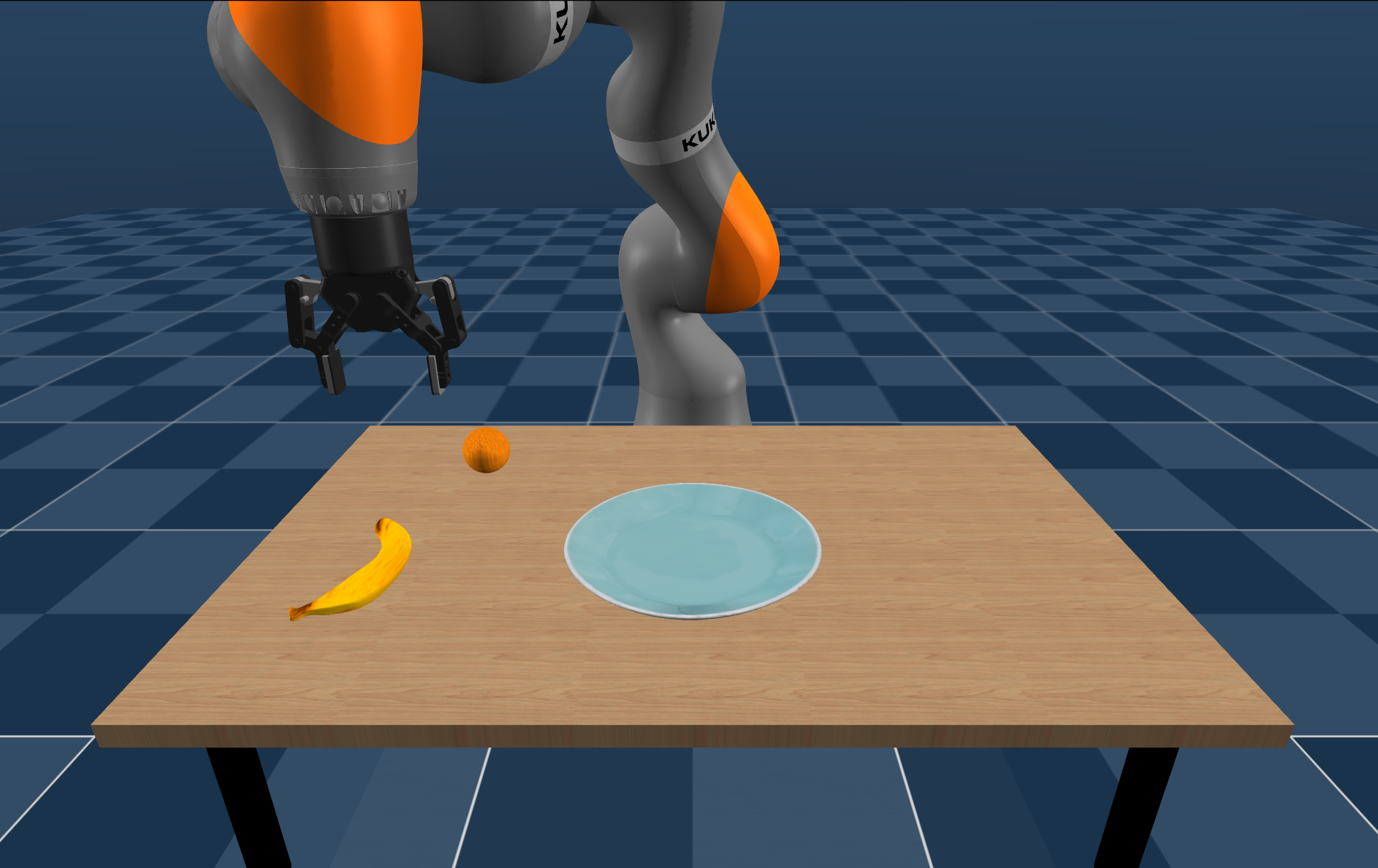}
\caption{Simulated fruit pick-and-place task in MuJoCo used for policy training experiment.}\label{fig:examples}
\end{figure}

\begin{figure}[t!]
\centering
\includegraphics[width=1.0\columnwidth]{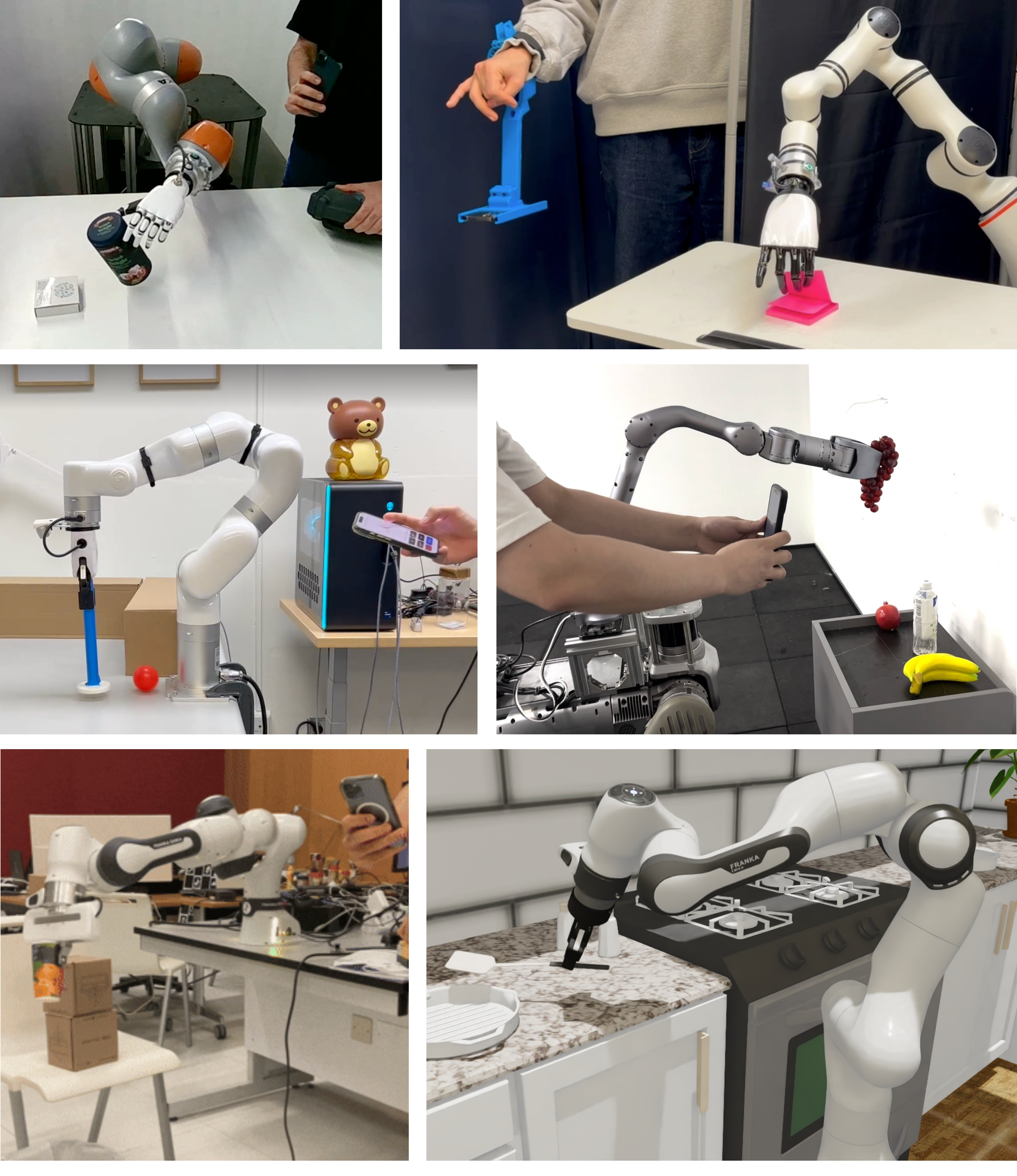}
\caption{Compilation of TeleDex deployments across multiple robot platforms in both our experiments and third-party settings.}
\label{fig:examples}
\end{figure}
% To verify that \Name{} demonstrations are suitable for learning, we construct a simple simulated pick-and-place task in which the robot must pick a banana and an orange and place them onto a plate. We collect 120 demonstration episodes and trained a simple baseline consisting of a vanilla CNN encoder followed by an MLP policy head. The resulting policy achieves a success rate of 80\% (16/20).
To verify that demonstrations collected with \Name{} are suitable for downstream policy learning, we design a simple simulated pick-and-place task in which the robot is required to sequentially grasp a banana and an orange and place them onto a plate. 

Using \Name{}, we collect a total of 120 teleoperated demonstration episodes. The observations consist of RGB images and proprioceptive robot states, and actions correspond to end-effector control commands. We then train a straightforward behavioral cloning baseline consisting of a vanilla CNN encoder for visual feature extraction, followed by a multi-layer perceptron (MLP) policy head that predicts control actions. No task-specific architectural modifications or additional data augmentation techniques are employed.

The trained policy achieves a success rate of 80\% (16/20) over 20 evaluation episodes, demonstrating that demonstrations collected with \Name{} are of sufficient quality to enable effective imitation learning, even with a simple baseline model and a relatively modest dataset size.

\subsection{Third-party Evaluations}

Beyond the evaluations presented in this paper, \Name{} has been used in several recent works including~\cite{gupta2025umi, lu2025gwm, zhi2025learning, kim2026molmospaces}, where it serves as a teleoperation interface for collecting robot manipulation data in both simulation and real-world settings. Examples of these deployments are shown in Fig.~\ref{fig:examples}. This allowed us to verify the ease of deployment of \Name{} across robotic platforms, including KUKA iiwa 14, Franka Panda, xArm-7, SO-100, RealMan, B2Z1, and the fleet of robots available in MolmoSpaces.

\section{Conclusion}
\label{sec:conclusion}

We presented \Name{}, an open-source system for accessible dexterous teleoperation using commodity smartphones. By streaming 6-DoF wrist pose and 21-DoF articulated hand state estimates without requiring external tracking infrastructure, \Name{} enables intuitive and portable control of robot arms and multi-fingered hands. The system combines a lightweight mobile application with a flexible Python API, supporting rapid deployment across diverse robotic platforms.

\smallskip
Through both quantitative and qualitative evaluations, we demonstrated that \Name{} facilitates efficient demonstration collection and supports precise, reactive manipulation in dynamic and contact-sensitive tasks. In addition, we showed that demonstrations collected with \Name{} are suitable for downstream policy learning, and that the system generalizes across multiple real-world robotic platforms. Together, these results highlight \Name{} as a practical and scalable solution for dexterous teleoperation and data collection.

\smallskip
\textbf{Limitations.} Although \Name{} substantially lowers the hardware and setup barriers for dexterous teleoperation, prolonged use of the wrist-mounted configuration may lead to user fatigue due to the added weight of the smartphone. This limitation is comparable to other wearable teleoperation devices, such as VR controllers, which similarly introduce ergonomic considerations during extended operation. Future work may explore lighter hardware configurations, alternative mounting strategies, or shared-autonomy approaches to further reduce operator burden.

\small
\bibliographystyle{IEEEtranN}
\bibliography{example}

\end{document}